\documentclass[runningheads]{llncs}
\usepackage[T1]{fontenc}
\usepackage{esvect}
\usepackage{graphicx}
\usepackage{hyperref}
\usepackage{enumitem}
\usepackage{amsmath}
\usepackage{amssymb}
\usepackage{booktabs}
\usepackage{caption}

\usepackage{subcaption}
\usepackage{adjustbox}
\usepackage{wrapfig}

\usepackage{diagbox}
\usepackage{pgfplots}
\usepackage{tikz}
\usetikzlibrary{shapes, arrows, positioning}
\pgfplotsset{compat=1.17}
\usepackage{multirow}

\usepackage{xcolor}
\usepackage{algorithm}
\usepackage{algorithmic}

\usepackage{svg}
\setsvg{inkscapeversion=1.2}

\begin{document}
\title{MC-GNNAS-Dock: Multi-criteria GNN-based Algorithm Selection for Molecular Docking}
\titlerunning{Multi-criteria GNN-based Algorithm Selection for Molecular Docking}
%
\author{Siyuan Cao\orcidID{0009-0002-1148-4381
} \and
Hongxuan Wu\orcidID{0009-0009-6499-7018} \and
Jiabao Brad Wang\orcidID{0009-0002-9559-6240} \and
Yiliang Yuan\orcidID{0009-0005-7266-7386} \and
Mustafa Misir\orcidID{0000-0002-6885-6775}}
  
\authorrunning{Cao et al.}

\institute{Duke Kunshan University, Kunshan 215316, China
\\
\email{\{siyuan.cao,hongxuan.wu,jb.wang,yiliang.yuan,mustafa.misir\}@dukekunshan.edu.cn}}

\maketitle              
\begin{abstract}
Molecular docking is a core tool in drug discovery for predicting ligand–target interactions. 
Despite the availability of diverse search-based and machine learning approaches, no single docking algorithm consistently dominates, as performance varies by context. To overcome this challenge, algorithm selection frameworks such as GNNAS-Dock~\cite{yuan2024gnnas}, built on graph neural networks, have been proposed. 
This study introduces an enhanced system, MC-GNNAS-Dock, with three key advances. 
First, a multi-criteria evaluation integrates binding-pose accuracy (RMSD) with validity checks from PoseBusters~\cite{buttenschoen2024posebusters}, offering a more rigorous assessment. 
Second, architectural refinements by inclusion of residual connections strengthen predictive robustness. Third, rank-aware loss functions are incorporated to sharpen rank learning. 
Extensive experiments are performed on a curated dataset containing $\sim$3200 protein-ligand complexes from PDBBind~\cite{liu2017forging}.
MC-GNNAS-Dock demonstrates consistently superior performance, achieving up to 5.4\%(3.4\%) gains under composite criteria of RMSD below 1~\AA(2~\AA) with PoseBuster-validity compared to the single best solver (SBS) Uni-Mol Docking V2~\cite{alcaide2024uni}. 

\keywords{algorithm selection, molecular docking, learning to rank}
\end{abstract}

\section{Introduction}
\label{sec:intro}

Molecular docking is a foundational computational technique in structure-based drug discovery, aiming to predict how small molecules (ligands) interact with target receptors (proteins) at an atomic level~\cite{fan2019progress}.
Docking methods encompass traditional ones which typically rely on heuristic searching to explore binding poses~\cite{fan2019progress}, and machine learning-based ones which directly predict those poses~\cite{cao2025surfdock}. However, in line with the \textit{No Free Lunch Theorem} (NFL)~\cite{wolpert2002no}, no single method consistently excels over all docking scenarios~\cite{yuan2024gnnas}. 

Algorithm Selection (AS) \cite{kerschke2019automated} traditionally refers to a system which predicts the most suitable algorithm based on instance-specific features.
Regarding docking, it was recently shown that an AS system using Graph Neural Networks (GNNs) to encode molecular graph features, \textit{GNNAS-Dock}~\cite{yuan2024gnnas}, can substantially outperform all standalone docking methods.
GNNAS-Dock relies on \textit{root mean square deviation} (RMSD)--geometric proximity to a reference pose--as its sole evaluation criterion. Nevertheless, it has recently been proven via \textit{PoseBusters}~\cite{buttenschoen2024posebusters},  a pose validation tool, that high-RMSD poses may still exhibit steric clashes or unrealistic geometries~\cite{alcaide2024uni,cao2025surfdock}. 
Moreover, GNNAS-Dock utilizes binary cross entropy (BCE) for training, overlooking the inherent affinity of AS with \textit{learning-to-rank} (LTR) techniques~\cite{misir2017alors}. 
Ranking-aware loss criteria such as Normalized Discounted Cumulative Gain (NDCG)~\cite{jarvelin2002cumulated} have demonstrated effectiveness in practical AS systems~\cite{chen2023algorithm}.

This paper proposes \textbf{MC-GNNAS-Dock}, extending GNNAS-Dock with multi-criteria evaluation that simultaneously accounts for positional accuracy, pose validity, and ranking quality. 
Our contributions are:
\begin{itemize}
    \item A scoring function integrating geometric accuracy (RMSD) and physical validity (PoseBusters) is introduced for algorithm evaluation.
(Sec.~\ref{subsec:scoring_func}).
    \item A decoder model embedding residual connections is adopted for feature-based algorithm selections (Sec.~\ref{subsec:model_archi}). 
    \item A composite loss function incorporating ranking-aware components emphasizing pairwise ranking (pairwise logistic loss) and top-candidate positions (NDCG loss) is proposed (Sec.~\ref{subsec:loss}).
    \item The proposed approaches are extensively evaluated on a selection of diverse ligand–protein complexes from PDBBind~\cite{liu2017forging} (Sec.~\ref{sec:results}).
\end{itemize}

\section{Background}
\label{sec:related-work}

\subsection{Algorithm selection} 

\paragraph{Definition.} Let $\mathcal{X}$ denote a set of problem instances and $\mathcal{A} = \{A_1, \dots, A_m\}$ a finite set of candidate algorithms. A general AS system is a mapping $S : \mathcal{X} \to \mathcal{A}$ that selects, for each $\mathbf x \in \mathcal{X}$, an algorithm expected to perform well under a task-specific evaluation score $\phi : \mathcal{A} \times \mathcal{X} \to \mathbb{R}_{\geq 0}$. 
In practice, $\phi$ is unknown and approximated by a learned model $\hat{\phi} : \mathcal{A} \times \mathcal{X} \to \mathbb{R}_{\geq 0}$. The system then defines
\begin{equation}
    S(x) := \arg\max_{A \in \mathcal{A}} \hat{\phi}(A, \mathbf x),
\end{equation}
thereby inducing selection via performance prediction. Usually, the model is in the form $\hat{\phi}(A, \mathbf x)  =  f(g(\mathbf x))\mid_A$ for a feature extractor (encoder) $g : \mathcal{X} \to \mathbb{R}^d$ and performance predictor (decoder) $f:\mathbb R^d \rightarrow \mathbb R^m$.

\paragraph{Algorithm selection for docking.} 
The first AS system specialized for molecular docking was proposed in~\cite{chen2023algorithm}, where an \textit{Algorithm Recommender System} (ALORS)~\cite{misir2017alors} is applied to configure an algorithm based on molecular features, outperforming any fixed choice. 
GNNAS-Dock~\cite{yuan2024gnnas} learns graph embeddings from the molecular graphs of proteins and ligands with GNNs to perform selection over a portfolio of advanced algorithms, outperforming all standalone methods.

\subsection{GNNAS-Dock}
\label{subsec:gnnas-dock}
\paragraph{Pipeline.} 
GNNAS-Dock models both ligands and proteins as graphs. \textit{Ligand graphs} $\mathcal G_L = (\mathcal V_L, \mathcal E_L)$ are constructed at an atomic level, with nodes (atoms) encoding atomic features (e.g., coordinates, atom types, atomic mass, etc.) and edges (chemical bonds) encoding bond types. \textit{Protein graphs} $\mathcal G_P = (\mathcal V_P, \mathcal E_P)$ are constructed at a residue level to manage computational complexity and scalability, with node (amino acid) features including centroidal coordinates, residue types, mass, etc., and edge (peptide bond) ones encoding bond lengths. 

The architecture follows a dual-encoder design, where a \textit{graphLambda}~\cite{mqawass2024graphlambda} model is used to process protein graphs to capture their multi-scale features, and a \textit{Graph Attention Network} (GAT)~\cite{velickovic2017graph} is used for ligand graphs. 

The ligand and protein embeddings are then concatenated and passed into a single-hidden-layer multi-layer perceptron (MLP), which predicts the performance scores of nine predefined docking algorithms, representative of state-of-the-art in their own types. This encoder-decoder pipeline is trained end-to-end to minimize the \textit{binary cross entropy} (BCE) in RMSD, with the labels obtained from docking experiments over a curated set of protein–ligand complexes.

\paragraph{Docking evaluation scores and optimization objectives.}
GNNAS-Dock optimizes based solely on RMSD, while PoseBusters~\cite{buttenschoen2024posebusters} highlights potential failure modes in low-RMSD poses, which has made PoseBuster-validity an increasingly standard performance metric in subsequent studies~\cite{alcaide2024uni,cao2025surfdock}.
Ranking optimization frameworks like \textbf{LambdaLoss}~\cite{wang2018lambdaloss} enables differentiable surrogates of ranking-aware metrics, having prompted LTR methods, but remain underutilized in algorithm selection though potential effectiveness is shown~\cite{misir2017alors}. Our work extends GNNAS-Dock by embedding multi-metric evaluation and ranking-aware penalities.

\section{Multi-Criteria GNNAS-Dock (MC-GNNAS-Dock)}
\label{sec:methodologies}

In this section, we introduce MC-GNNAS-Dock, which incorporates a composite scoring function encoding both geometric accuracy and physical feasibility, and a refactored loss function integrating ranking-aware penalties. 

\subsection{Evaluation score}
\label{subsec:scoring_func}
\paragraph{Geometric accuracy via RMSD.}
RMSD is the standard metric for quantifying the spatial error of a predicted pose with respect to its crystallographic reference. In docking, a pose is usually deemed \emph{near-native} if $\text{RMSD}\leq 2$~\AA\ and \emph{irrelevant} if $\text{RMSD}>5$~\AA. Since minor sub-\AA\ improvements yield diminishing interpretability once plausibility is established, we adopt a \textit{normalized}, monotonically decreasing \textit{exponential} scoring function clipped at upper tolerance $M$:
\begin{equation}\label{eq:rmsd_score}
  s_{\text{RMSD}}(x) \ =\
  \begin{cases}
    \displaystyle\frac{e^{M}-e^{x}}{e^{M}-1} & x\le M,\\[6pt]
    0 & x>M,
  \end{cases}
\end{equation}where $x$ is the RMSD in~\AA.
Thus $s_{\text{RMSD}}(0)=1$ (perfect alignment), $s_{\text{RMSD}}(M)=0$ (geometric failure), and the score decreases smoothly in between. 

\paragraph{Chemical Validity via PoseBusters.}
We apply a strict accept–reject criterion that mirrors current practice in crystallographic validation:
\begin{equation}\label{eq:pb_score}
   s_{\text{PB}}
  \ =\
  \begin{cases}
    1 \quad & \text{pass all $18$ PoseBusters checks (i.e., \textit{PB-valid}),} \\
    0 & \text{otherwise,}
  \end{cases} 
\end{equation}
which ensures that no geometrically plausible yet chemically impossible pose receives a decent score. We propose two versions of the composite score, in particular with \textbf{additive combination} ($s_{\alpha} = \alpha s_{\text{RMSD}} + (1-\alpha)s_{\text{PB}},\ \alpha\in[0,1]$) which accounts for individual contribution of each and \textbf{multiplicative combination} ($s_{\text{mul}} = s_{\text{RMSD}} \cdot s_{\text{PB}}$) which simulated a RMSD score gated by PB-validity, with an ablation carried in Sec.~\ref{subsec:setup}.

\subsection{Ranking-aware loss function}
\label{subsec:loss}

\paragraph{Pairwise Logistic (PL) Loss.} 
PL loss recasts the ranking task as a binary classification over pairs. Given a pair of score labels $(y_i, y_j)$ and predicted scores \( (s_i, s_j) = (\phi(A_i, \mathbf x), \phi(A_j, \mathbf x))\), the Bradley-Terry model is used to estimate the probability that item \( i \) should be ranked above \( j \):
\begin{equation}\label{eq:BT_basic}
    \mathbb P(y_i>y_j|s_{i\mid j}) = \left(1 + e^{-\sigma(s_i - s_j)}\right)^{-1},
\end{equation}
with a temperature-like scaling constant \( \sigma \), for which the conventional value $1$ is adopted. 
The loss then takes the form of the negated log-likelihood:
\begin{equation}
    \mathcal{L}_{\text{PL}}(i, j) = \log \left(1 + e^{-\sigma(s_i - s_j)} \right).
\end{equation}
This pairwise formulation provides a smooth, differentiable surrogate for equally treating the orders of pairs. In addition, we recognize top-position attention can also be advantageous for selecting algorithms over a diverse portfolio. We hereby also incorporate a position-sensitive measure to our loss.

\paragraph{NDCG-Loss2.} NDCG is a family of ranking optimization metrics characterized in emphasizing top-positioned relevant items. For $n$ items, it takes the form
\begin{equation}\label{eq:NDCG}
    \text{NDCG} = \sum_{i=1}^n\frac{G_i}{D_i}, 
\end{equation}
where $G_i$ denotes the \textit{gain} from assigning relevance label $y_i$ to position $i$, and $D_i$ is a \textit{discount} factor penalizing lower-ranked positions. 
A differentiable and computationally light surrogate, \textit{NDCG-Loss2}~\cite{wang2018lambdaloss}, is constructed via marginal inference on the likelihood of observing $y_i > y_j$ given the scores over all rankings $\Pi$, $ \mathbb P(y_i>y_j|s_{i\mid j}) = \sum_{\pi\in\Pi} \mathbb P(y_i>y_j|s_{i\mid j}, \pi_{i\mid j})\mathbb P(\pi|\mathbf s)$ for $\mathbf s = \{s_i\}_{i=1,...,m}$, 
and deploying a rank-sensitive version of Eq~\ref{eq:BT_basic} 
\begin{equation}\label{eq:BT_ndcg}
    \mathbb P\left(y_i>y_j| s_{i\mid j}, \pi_{i\mid j}\right) = \left(1 + e^{-\sigma(s_i - s_j)}\right)^{-|G_i - G_j|\cdot\left|D_{\pi_i}^{-1} - D_{\pi_j}^{-1}\right|}, 
\end{equation}
where the power measures how ranking performance varies if $i, j$ are switched in ranking $\pi$. For computational efficiency, one can compute the (negated) data-based expected log-likelihood:
\begin{equation}
    l_D\left(\mathbf y, \mathbf s\right) \ =\ - \sum_{\pi\in\Pi} \mathbb P(\pi| \mathbf s)\log_2\mathbb P\left(y_i>y_j| s_{i\mid j}, \pi_{i\mid j}\right).
\end{equation}
The total loss function is thus given by the mean over all samples:
\begin{equation}
    \mathcal L_{\text{NDCG}}\left(\hat\phi\right) = \frac{1}{n} \sum_{(\mathbf{y}, \mathbf{x})\in (\mathcal Y, \mathcal X)} l_D\left(\mathbf y, \mathbf s\right).
\end{equation}

\subsection{Model architecture}
\label{subsec:model_archi}
MC-GNNAS-Dock adopts the setup of protein/ligand graph embeddings and the GNN encoder architectures from GNNAS-Dock. For the decoder, a single-hidden-layer MLP is of limited representativeness for handling multi-criteria evaluation. Inspired by the generalizability of ResNet~\cite{he2020resnet}, we implement a 2-hidden-layer MLP with 2 sets of residual block triplets with hidden dims 256/128 respectively. The outputs from the residual blocks are concatenated and fed through a linear layer for dimension reduction. Our implementation is available at \url{https://github.com/ToothlessOS/MC-GNNAS-Dock.}

\section{Experiments}
\label{sec:results}

\subsection{Setup}\label{subsec:setup}

\paragraph{Dataset and Preprocessing.} We evaluated our approach on a benchmark set derived from PDBBind, consisting of around 3200 protein-ligand complexes with diverse binding site characteristics and molecular topologies. 
All results in Sec.~\ref{subsec:results} are based on 10-fold cross-validations with 9-1 train-test splits, and statistical tests are performed against SBS and the sole BCE loss respectively. 

\paragraph{Baseline Methods. } 
MC-GNNAS-Dock is evaluated over a curated portfolio of eight traditional or ML-based docking algorithms, as summarized in Table~\ref{tab:docking_tools}. This selection reflects the observed trade-off: traditional methods tend to favor physicochemical plausibility, while ML-based ones often excel in RMSD performance~\cite{buttenschoen2024posebusters}. Each included method is either a state-of-the-art (SOTA) representative of its category or a widely used empirical baseline.
\begin{table}[ht!]
\vspace{-2em}
\centering
\caption{Candidate docking algorithms for selection}
\begin{tabular}{|l|l|l|}
\hline
\textbf{Method} & \textbf{Type} &\textbf{Central mechanism}\\
\hline
Smina~\cite{koes2013lessons} & Classical &Vina with empirical scoring and local search\\
\hline
Qvina~\cite{hassan2017protein} & Classical &Vina with fast stochastic global optimization\\
\hline
DiffDock~\cite{corso2022diffdock}& ML-based &Diffusion model for direct pose generation\\\hline
DiffDock-L~\cite{corso2024deep}& ML-based &DiffDock with flexible ligand modeling\\
\hline
SurfDock~\cite{cao2025surfdock} & ML-based &$SE(3)$-equivariant diffusion GNN\\
\hline
Gnina~\cite{mcnutt2021gnina} & ML-based &$3$D CNN-based rescoring of docking poses\\
\hline
Uni-Mol (Docking V2)~\cite{alcaide2024uni} & ML-based &Transformer on atomic coordinates with pretraining\\
\hline
KarmaDock~\cite{zhang2023efficient} & ML-based &$SE(3)$-equivariant attention\\
\hline
\end{tabular}
\vspace{-1.5em}
\label{tab:docking_tools}
\end{table}

\paragraph{Experimental Design.} 
Extensive experiments are performed over different model configurations including decoder structures and loss components, encompassing combinations of:
\begin{itemize}
    \item Decoder: The residual MLP described in Sec.~\ref{subsec:model_archi} (\texttt{residual}); Same MLP without residual connections (\texttt{MLP}).
    \item Loss: \textbf{BCE} only (\texttt{BCE}); \textbf{BCE} + \textbf{NDCG} (\texttt{+NDCG}); \textbf{BCE} + \textbf{PL} (\texttt{+PL}); \textbf{BCE} + \textbf{PL} + \textbf{NDCG} (\texttt{+Both}).
\end{itemize}
In addition, excessive algorithm portfolio size can dilute learning signal and increase model complexity without proportional gains~\cite{kerschke2019automated}, especially when some algorithms consistently underperform or show high redundancy through correlated behaviors. To mitigate this, we also conduct mirror experiments using a refined portfolio consisting of the top-4 performing algorithms on training data (Qvina, SurfDock, Gnina, and Uni-Mol).

\paragraph{Score ablation.} An ablation study on evaluation scores (Sec.~\ref{subsec:scoring_func}) is conducted using the baseline (\texttt{residual decoder} + \texttt{BCE}) with a fixed train/test split due to the combinatorial number of configurations. The tested $M$ ranges from $1$ denoting exquisite geometric precision to $5$ indicating invalid spatial accuracy. The results are shown in Table~\ref{tab:score_ablation1} and ~\ref{tab:score_ablation2}. 
The integer-$M$ rows suggest an optimal range of trade-off in $[2, 3]$. We optionally parametrize $M=\ln n$ to interpret the exponential range $e^M$ in terms of discrete tolerable deviations, and a finer research identifies $s_{\text{mul}}$ with $M=\ln 11\approx 2.40$ as optimal.
\begin{figure}[ht]
\vspace{-1.5em}
  \centering
  \adjustbox{valign=t}{%
    \begin{minipage}[t]{0.48\textwidth}
      \centering
      \captionof{table}{(\%) $\text{RMSD}\leq1\text{\AA}$ \& PB-valid}
      \begin{tabular}{c|ccccc|c}\toprule
        \diagbox{$M$}{$\alpha$}& $0.1$&$ 0.3$& $0.5$&$0.7$ &$0.9 $ &$s_{\text{mul}}$\\
        \hline
 $1$& 25.79& 22.33&43.08&42.14& 44.97 &45.28\\
 $2$& 22.01& 48.11& 48.11&47.48& 46.23 &51.26\\
 $\ln 11$& 38.68& 45.28& 49.37& 50.31&52.52 &\textbf{54.72}\\
 $3$& 26.73& 33.65&45.60&46.54& 41.07 &45.91\\
        $5$& 22.01& 35.85& 44.03&40.88& 43.71 &47.17\\
        \bottomrule
      \end{tabular}
      \label{tab:score_ablation1}
    \end{minipage}
  }
  \hspace{1pt}
  \hfill
  \adjustbox{valign=t}{%
    \begin{minipage}[t]{0.48\textwidth}
      \centering
      \captionof{table}{(\%) $\text{RMSD}\leq2\text{\AA}$ \& PB-valid}
      \begin{tabular}{c|ccccc|c}\toprule
        \diagbox{$M$}{$\alpha$}& $0.1$& $0.3$& $0.5$&$0.7$ &$0.9$  &$s_{\text{mul}}$\\
        \hline
 $1$& 40.57& 34.28&59.12&54.72& 63.21 &60.06\\
 $2$& 34.28& 70.75& 70.07&67.61& 66.98 &71.38\\
 $\ln 11$& 58.49& 67.3& 72.96& 72.33&72.01 &\textbf{76.42}\\
 $3$& 42.45& 48.43&69.50&68.87& 49.37 &69.81\\
        $5$& 33.65& 53.46& 63.84&61.01& 62.26 &72.96\\
        \bottomrule  
      \end{tabular}
      \label{tab:score_ablation2}
    \end{minipage}
  }
\vspace{-1.5em}
\end{figure}
\vspace{-1em}

\paragraph{Implementation.} The graph embeddings of ligands and proteins are obtained using Python package \texttt{Graphein}~\cite{jamasb2022graphein}. 
All experiments were conducted in \texttt{Python} on a dual-socket server equipped with two AMD EPYC 9654 CPUs and 8 NVIDIA GeForce RTX 4090 GPUs, running \texttt{Ubuntu 22.04.1 LTS}. On average, it takes approximately 5 seconds to build graphs for a protein-ligand pair and around 0.5 seconds for single inference. 

\subsection{Results}\label{subsec:results}
The computational results are reported in Table~\ref{tab:scores}. 
Across all settings, MC-GNNAS-Dock consistently outperforms the SBS baseline (Uni-Mol), demonstrating gains ranging \textbf{3.3\%-5.4\%} and \textbf{2.2\%-3.4\%} over SBS in the 1~\AA\ \& PB-valid and 2~\AA\ \& PB-valid gated percentages, all statistically significant. The best model (\texttt{Residual+BCE}) achieves around \textbf{48.8\%} and \textbf{71.6\%} absolute values against SBS's \textbf{43.4\%} and \textbf{68.2\%} in those two measures.
\begin{table}[ht!]
\vspace{-2em}
\centering
\caption{Averaged 10-fold results are in \%. $\Delta$ is absolute gain over SBS. Significance: “*” denotes $p<0.05$ v.s. SBS (paired test across tasks) and \textbf{LTR} represents $p<0.05$ v.s. \texttt{BCE}. 
The best results and SBS performance are in \textbf{bold}.
}
\label{tab:scores}
\setlength{\tabcolsep}{6pt}
\begin{tabular}{lcc@{\;}lcc@{\;}lc}
\toprule
\textbf{Method} & 
\multicolumn{2}{c}{$\leq 1\,\text{\AA}\ \&\ \text{PB-valid}$} &
& \multicolumn{2}{c}{$\leq 2\,\text{\AA}\ \&\ \text{PB-valid}$} &
& \textbf{LTR} \\
\cmidrule(lr){2-3}\cmidrule(lr){5-6}
 & \multicolumn{1}{c}{Abs} & \multicolumn{1}{c}{$\Delta$} && \multicolumn{1}{c}{Abs} & \multicolumn{1}{c}{$\Delta$} && \\
\midrule
 Smina     & 24.4 &   --   && 38.6 &   --   && -- \\
 Qvina     & 24.3 &   --   && 38.0 &   --   && -- \\
\cmidrule(lr){1-8}
 KarmaDock &  3.1 &   --   &&  8.5 &   --   && -- \\
 DiffDock   & 8.6 &   --   && 11.4 &   --   && -- \\
 DiffDock-L & 10.2 &   --   && 14.7 &   --   && -- \\
 Gnina     & 25.4 &   --   && 42.9 &   --   && -- \\
 SurfDock  & 39.3 &   --   && 53.7 &   --   && -- \\
 Uni-Mol (SBS) & \textbf{43.4} & \textbf{--} && \textbf{68.2} & \textbf{--} && -- \\
\addlinespace[2pt]
\multicolumn{8}{l}{\emph{MC-GNNAS-Dock (4 algorithms)}}\\
\cmidrule(lr){1-8}
 \texttt{MLP (BCE)}            & 47.0 & +3.6* && 71.0 & +2.8* && -- \\
 \texttt{MLP (+NDCG)}          & 47.7 & +4.3* && 70.8 & +2.6* && No \\
 \texttt{MLP (+PL)}            & 47.0 & +3.6* && 70.5 & +2.3* && No \\
 \texttt{MLP (+Both)}          & 47.8 & +4.4* && 70.5 & +2.3* && No \\
\cmidrule(lr){1-8}
 \texttt{Residual (BCE)}       & 48.3 & +4.9* && 71.1 & +2.9* && -- \\
 \texttt{Residual (+NDCG)}     & 48.3 & +4.9* && 71.2 & +3.1* && No \\
 \texttt{Residual (+PL)}       & 48.5 & +5.1* && 71.0 & +2.8* && No \\
 \texttt{Residual (+Both)}     & 48.1 & +4.7* && 70.8 & +2.6* && No \\
\addlinespace[2pt]
\multicolumn{8}{l}{\emph{MC-GNNAS-Dock (8 algorithms)}}\\
\cmidrule(lr){1-8}
 \texttt{MLP (BCE)}            & 46.7 & +3.3* && 70.5 & +2.3* && -- \\
 \texttt{MLP (+NDCG)}          & 47.6 & +4.2* && 70.6 & +2.4* && No \\
 \texttt{MLP (+PL)}            & 48.0 & +4.6* && 70.4 & +2.2* && No \\
 \texttt{MLP (+Both)}          & 48.0 & +4.6* && 71.4 & +3.2* && \textbf{Yes} \\
\cmidrule(lr){1-8}
 \texttt{Residual (BCE)}       & \textbf{48.8} & \textbf{+5.4}* && \textbf{71.6} & \textbf{+3.4}* && -- \\
 \texttt{Residual (+NDCG)}     & \textbf{48.8} & \textbf{+5.4}* && 71.3 & +3.1* && No \\
 \texttt{Residual (+PL)}       & \textbf{48.8} & \textbf{+5.4}* && 71.1 & +2.9* && No \\
 \texttt{Residual (+Both)}     & 48.6 & +5.2* && 71.0 & +2.9* && No \\
\bottomrule
\end{tabular}
\vspace{-2em}
\end{table}

\paragraph{Loss function effects.} The effect of incorporating ranking-aware loss terms NDCG and PL is \emph{not} uniformly beneficial, but context-dependent. In several \texttt{MLP} variants, especially with larger portfolios, the combined use of BCE with both ranking terms leads to clear gains, e.g., \texttt{MLP+Both} (48.0\%, 71.4\%) v.s. \texttt{MLP+BCE} (46.7\%, 70.5\%) with 8 algorithms. However, in \texttt{residual} cases, these additions offer marginal or no improvement over \texttt{BCE}. Furthermore, the inclusion of NDCG and PL yields comparable performance, with no clear advantage observed between the two. These suggest that ranking-aware signals can enrich training by aligning the model's output with ordinal preferences, but may also introduce conflicting gradients or redundant supervision, particularly when the base loss already yields strong alignment. Their effectiveness may depend on architectural capacity, loss weighting, and portfolio diversity, inviting further exploration.

\paragraph{Architectural trends.} Residual decoders systematically outperform MLP decoders, with gains of up to 2.1\% at the 1~\AA\ level and 1.1\% at the 2~\AA\ level. These improvements are consistent and suggest that residual connections support better gradient propagation and feature reuse, suitable for molecular graph embeddings.

\begin{figure}[ht]
\vspace{-1.5em}
\centering
\begin{tikzpicture}
\begin{axis}[
    height=6cm,
    width=12cm,
    ybar=0pt,
    bar width=7pt,
    enlarge x limits=0.1,
    enlarge y limits=0.15,
    ylabel={Selection Frequency},
    symbolic x coords={Uni-Mol, SurfDock, Smina, Gnina, DiffDock-L, Qvina, DiffDock, KarmaDock},
    xtick=data,
    x tick label style={rotate=45, anchor=east},
    legend style={at={(0.5,0.8)}, anchor=south, legend columns=-1},
    nodes near coords,
    ymin=0,
    ymajorgrids=true,
    grid style=dashed
]

\addplot+[
    fill=cyan!50,
    nodes near coords,
    every node near coord/.append style={color=cyan!70}
] coordinates {
    (Uni-Mol,892) (SurfDock,1222) (Smina,225) (Gnina,354)
    (DiffDock-L,139) (Qvina,209) (DiffDock,118) (KarmaDock,20)
};

\addplot+[
    fill=purple!70,
    nodes near coords,
    every node near coord/.append style={color=purple!90}
] coordinates {
    (Uni-Mol,1806) (SurfDock,696) (Smina,420) (Gnina,168)
    (DiffDock-L,54) (Qvina,28) (DiffDock,4) (KarmaDock,3)
};

\legend{VBS, \texttt{Residual} (\texttt{BCE})}
\end{axis}
\end{tikzpicture}
\caption{Selection frequencies under VBS and \texttt{residual} (\texttt{BCE}) (8 algorithms).}
\label{fig:frequencies}
\vspace{-1.5em}
\end{figure}
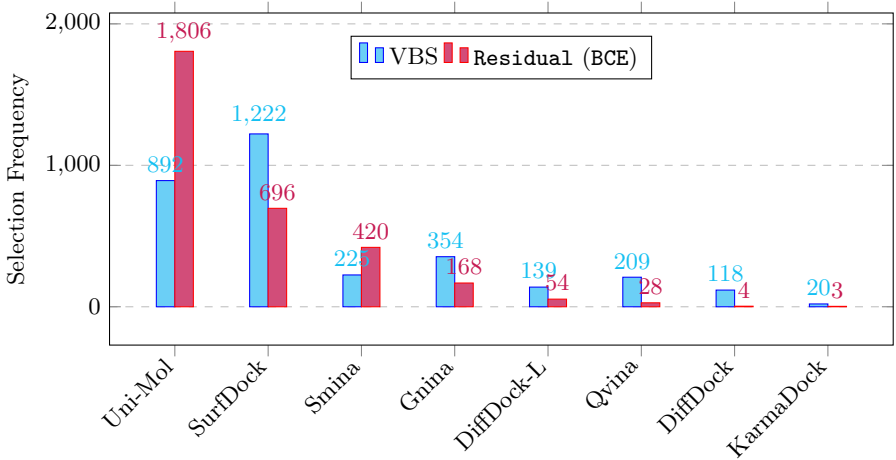

\paragraph{Scalability to larger portfolios.} The model consistently achieves \emph{stronger} performance when scaled to a more diverse 8-algorithm portfolio. The selection frequencies over the 10 folds of the virtual best solver (VBS), which selects the best algorithm for each instance, and our best model, \texttt{Residual} (\texttt{BCE}), are shown in Fig.~\ref{fig:frequencies}. VBS selection involves all 8 algorithms and best results are observed in this setting, confirming MC-GNNAS-Dock’s ability to exploit performance complementarity across methods. 

\paragraph{Score-induced preference.} Compared to SurfDock, Uni-Mol exhibits a slightly higher \emph{lower-quartile} RMSD (0.70~\AA\ v.s. 0.56~\AA) but a lower \emph{mean} (1.31~\AA\ v.s. 1.58~\AA) and higher valid percentage (68.2\% v.s. 53.7\%), indicating fewer large-error tails. Consequently, our selector favors Uni-Mol’s robustness, whereas VBS more often chooses SurfDock due to its numerous marginal wins. This behavior is consistent with the exponential score, which saturates near-zero RMSD, de-emphasizing sub-\AA\ differences while assigning larger gains in the near-2~\AA\ region.

\section{Conclusion}
\label{sec:discussion}
\paragraph{MC-GNNAS-Dock.} We have extended GNN-based algorithm selection system, GNNAS-Dock, with multi-criteria evaluation. By integrating spatial accuracy, structural validity, and ranking-aware penalties, our method balances geometric plausibility with biochemical correctness, and yields improved consistency across diverse docking scenarios. 

\paragraph{Future work.} 
This study primarily aims to demonstrate the promise of GNN-based algorithm selection with multi-criteria evaluation. To maintain focus, we adopt \textit{default} settings for the ranking-aware loss terms, without exploring their parameter sensitivity or interpretability. Future work could refine performance and test robustness by tuning internal parameters and weighting strategies.
Moreover, extending evaluations beyond \textit{self-docking} scenarios with traditional and ML-docking methods, e.g. to \textit{cross-docking} cases and ML-based \textit{co-folding} methods like in \textit{PoseX}~\cite{jiang2025posex}, would further enhance generalizability.

\begin{credits}
\subsubsection{\ackname} The research results of this article are sponsored by the Wang-Cai Biochemistry Lab and Synear Food Molecular Biology Lab.
\end{credits}

%
%
\bibliographystyle{splncs04}
\bibliography{references}
\end{document}